\documentclass[runningheads]{llncs}
\usepackage[T1]{fontenc}
\usepackage{cite}
\usepackage{amsmath,amssymb,amsfonts}
\usepackage{algorithm}
\usepackage{algorithmic}
\usepackage{graphicx}
\usepackage{textcomp}
\usepackage{xcolor}
\usepackage{booktabs}
\usepackage{subcaption}

\usepackage{makecell, array, adjustbox, xcolor, multirow}
\begin{document}
\title{Efficient Mathematical Reasoning Models via Dynamic Pruning and Knowledge Distillation}
\titlerunning{Efficient Math Reasoning via Pruning and Distillation}
%
\author{Fengming Yu, Qingyu Meng, Haiwei Pan, Kejia Zhang
}
\authorrunning{F. Yu et al.}
%
\institute{Harbin Engineering University, Harbin, China \\ 
\email{\{yufengming, qingyumeng, panhaiwei, kejiazhang\}@hrbeu.edu.cn}\\
}
\maketitle              
\begin{abstract}

With the rapid development of deep learning, large language models have shown strong capabilities in complex 
reasoning tasks such as mathematical equation solving. 
However, their substantial computational and storage costs hinder practical deployment. 
This paper proposes a lightweight optimization method that integrates dynamic attention head pruning with knowledge distillation. 
The approach dynamically evaluates the importance of each attention head in the multi-head attention mechanism using a 
combination of weight norms and entropy, and prunes redundant heads in real time to reduce computational overhead. 
To mitigate performance degradation, knowledge distillation transfers information from the original model to the pruned 
student, enabling the smaller model to preserve reasoning ability.
Experiments conducted on both Math23k and ASDiv-A verify the effectiveness of the proposed method. 
For example, on Math23k with a 30\% pruning ratio, parameters are reduced by 18.7\%, inference speed is improved by 
27.5\%, FLOPs are reduced by 19.3\%, and accuracy drops only 0.7\% (from 84.4\% to 83.7\%). 
These results demonstrate that the method achieves substantial efficiency gains while maintaining strong reasoning 
performance, providing a practical solution for efficient deployment of large language models in mathematical reasoning 
tasks.

\end{abstract}

\section{Introduction}

The Transformer architecture\cite{vaswani2017attention} has revolutionized the field of natural language processing (NLP) and 
has achieved outstanding results in mathematical reasoning tasks. Large Transformer-based language models, such as GPT-3,
ATHENA and MWP-BERT, have performed well in the task of constructing and solving systems of mathematical equations, 
significantly outperforming traditional methods\cite{brown2020language,kim2023athena,liang2021mwp,zong2023solving}. 
This success has driven the development of larger models, and the Transformer architecture has been extended to handle 
increasingly complex mathematical reasoning tasks. However, this expansion brings huge computational requirements, 
especially since the multi-head attention mechanism requires complex association calculations between each position in the sequence. 
These computational requirements, coupled with high inference costs and energy consumption, have created a huge obstacle to 
practical deployment, particularly in mobile, edge, and embedded environments where efficiency and limited resources are key concerns. 
For example, GPT-3\cite{brown2020language} has 175 billion parameters and requires massive computational resources even 
at the inference stage.

To meet the urgent need for more efficient and lightweight reasoning models, 
the research community has focused on optimizing the Transformer architecture. 
Numerous strategies have been proposed, including pruning, quantization, and knowledge distillation. 
Among these, optimizing the multi-head attention mechanism has emerged as a promising approach. 
Studies have shown that not all attention heads contribute substantially to reasoning performance, and some heads have 
little effect or may even introduce noise\cite{jin2024moh}. 
This provides a theoretical basis for achieving compression by pruning redundant attention heads.

Most existing optimization methods, especially for the attention mechanism, require comprehensive retraining. 
This process is a daunting challenge for models with large parameter counts, demanding significant computational 
resources and time. Such retraining is not only economically costly but also raises environmental concerns due to the 
energy consumption involved. This highlights the need for more efficient methods to adapt and optimize large-scale 
inference models for diverse deployment scenarios.

To address these challenges, a lightweight optimization method is proposed that integrates dynamic attention head pruning 
with knowledge distillation, specifically designed for mathematical reasoning tasks. 
The core idea is to dynamically evaluate the importance of each attention head during equation construction and solution, 
using a combination of attention weight norm and attention entropy as indicators. 
Redundant heads are pruned in real time, adaptively optimizing the model structure and reducing computational overhead. 
To mitigate potential performance loss, a recursive knowledge distillation framework is introduced. 
Initially, the unpruned model serves as a teacher to guide the training of a pruned student model, 
which is then iteratively refined and adopted as the new teacher in successive pruning-distillation cycles. 
By jointly leveraging dynamic pruning and recursive knowledge distillation, this progressive strategy enables effective 
compression while maintaining robust reasoning performance, and facilitates efficient deployment of large language 
models for mathematical reasoning in resource-limited settings, contributing to the broader goals of sustainable and 
pervasive computing.

\section{Related Work}
\label{related_work}

\subsection{Large Language Models in Mathematical Reasoning Tasks}

Mathematical reasoning has progressed from early rule-based systems relying on templates and feature engineering 
\cite{feigenbaum1963computers} to deep learning methods such as RNNs and Seq2Seq models that map natural language to 
mathematical expressions \cite{ling2017program,wang2017deep}. However, their limited ability to capture long-range 
dependencies hindered complex reasoning \cite{bengio1994learning}. The introduction of Transformers \cite{vaswani2017attention} 
and pre-trained language models like GPT-3 and MWP-BERT \cite{brown2020language,liang2021mwp} significantly improved 
performance by leveraging self-attention, while graph-based approaches further enhanced quantitative relation modeling \cite{zhang2020graph}. 
More recently, multimodal Transformers have emerged, integrating text and visual features. 
Despite these advances, large models remain resource-intensive, making efficiency and model compression critical 
challenges for practical deployment.

\subsection{Model Lightweighting Techniques}

Large language models excel at mathematical reasoning but their massive size hinders deployment in resource-limited 
settings \cite{brown2020language}. To address this, lightweight techniques such as pruning, quantization, and knowledge 
distillation have been explored \cite{hoefler2021sparsity,han2015learning,sanh2019distilbert,gou2021knowledge,2024-40694}. 
Pruning reduces redundant parameters, with dynamic strategies adapting to input complexity for better 
efficiency \cite{frankle2020lottery,gao2018dynamic,fang2023depgraph,frantar2023sparsegpt}, 
while distillation transfers knowledge from large teachers to compact students, 
leveraging soft outputs and tailored loss functions to preserve reasoning ability \cite{hinton2015distilling,DBLP:conf/iccv/SonNCH21,DBLP:journals/corr/RomeroBKCGB14,DBLP:conf/iccv/LaoSLLY23,DBLP:conf/cvpr/YimJBK17,DBLP:conf/iclr/XuFZXWDX022}. 
Although effective in vision and NLP, these methods face stricter demands in mathematical reasoning, where symbolic 
logic and numerical accuracy are critical. Thus, combining dynamic pruning and distillation offers a promising path to 
balance efficiency and performance for real-world deployment.

\section{Method}
\label{methodology}

\begin{figure*}[!htbp]
\centering
\includegraphics[width=\textwidth]{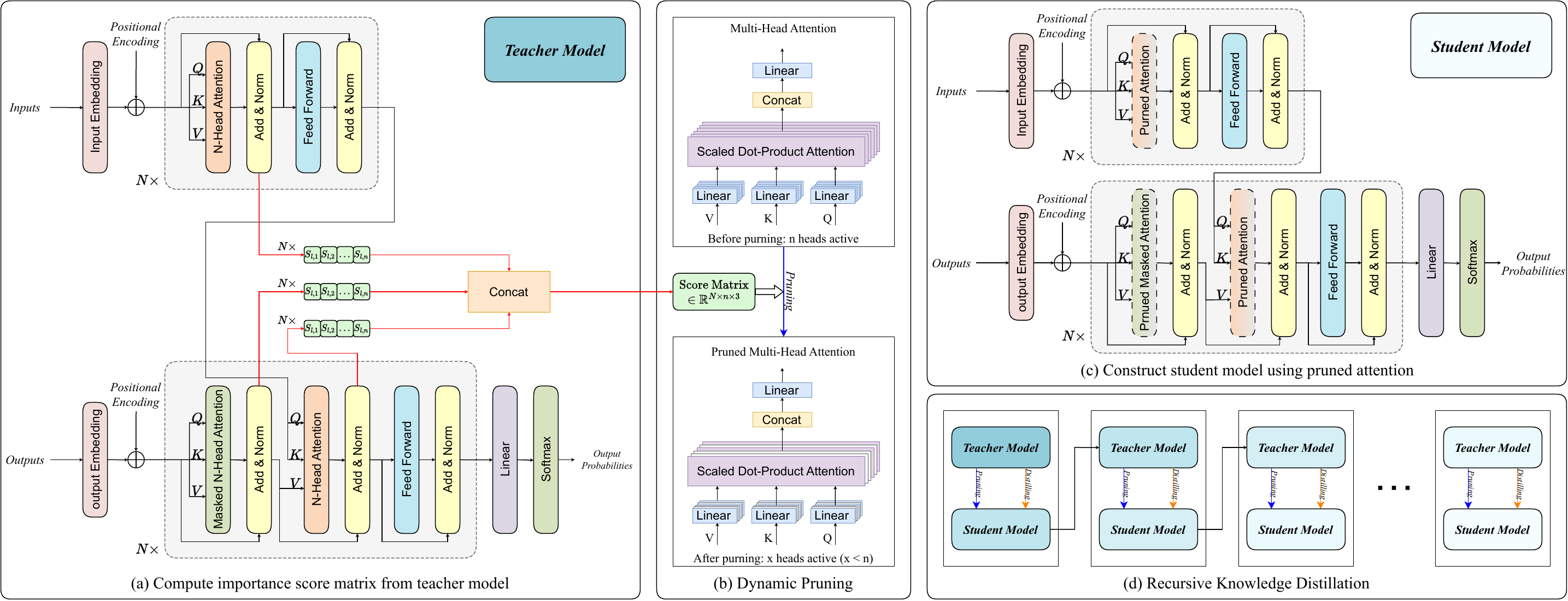}
\caption{Overview of the pruning and distillation framework. (a) The teacher model performs inference on the dataset 
to calculate the importance scores of each attention head, generating an importance score matrix; (b) Dynamic pruning 
based on importance score matrix; (c) The pruned model serves as student and is distilled from the unpruned teacher model.
(d) Recursive knowledge distillation: each student serves as the teacher for the next stage until the target compression 
ratio is reached.}
\label{fig:overall_architecture}
\end{figure*}

The proposed method integrates dynamic pruning with knowledge distillation to compress the model while preserving 
inference performance, as illustrated in Figure~\ref{fig:overall_architecture}. 
A dynamic pruning strategy evaluates the importance of each attention head using both weight norm and attention entropy. 
During training, redundant heads are pruned in real time, enabling adaptive structural optimization. 
For a layer with $n$ heads, a score vector of length $n$ is generated, and for a model with $N$ encoder and $N$ decoder 
layers, $3N$ such vectors are produced. Attention heads with lower scores are removed to achieve compression. 

To mitigate performance loss caused by pruning, a recursive knowledge distillation strategy is employed.
In the first stage the unpruned model serves as teacher for the pruned student, and in the second stage the student 
becomes the new teacher so that pruning and distillation proceed iteratively, progressively compressing the model while 
transferring knowledge.

\subsection{Evaluation of Attention Head Importance}

The importance of each attention head is evaluated by combining weight norm and attention entropy, 
which capture structural parameters and distribution characteristics respectively. 
For the $i$-th head in the $l$-th layer, the weight norm is defined as
\begin{equation}
w_{l,i} = \frac{1}{d_k} \sum_{j=1}^{d_k}(\left|Q_{l,i,j}\right| + \left|K_{l,i,j}\right| + \left|V_{l,i,j}\right|),
\label{eq:attention_weight_norm}
\end{equation}
reflecting the parameter magnitude and potential expressive capacity of the head.
The attention distribution is represented by the weight matrix
\begin{equation}
A_{l,i} = \mathrm{Softmax}\!\left(Q_{l,i} K_{l,i}^T / \sqrt{d_k}\right),
\end{equation}
and its entropy is defined as
\begin{equation}
H_{l,i} = -\sum_{j=1}^{d_k} A_{l,i,j} \log A_{l,i,j},
\label{eq:attention_entropy}
\end{equation}
where lower entropy indicates a more concentrated focus and higher entropy corresponds to a more uniform allocation.

The two indicators are integrated into a comprehensive importance score:
\begin{equation}
S_{l,i} = \alpha \cdot w_{l,i} + (1-\alpha) \cdot H_{l,i},
\label{eq:importance_score}
\end{equation}
with $\alpha \in [0,1]$ balancing their relative contributions. By adjusting $\alpha$, a flexible trade-off is achieved 
between parameter scale and distribution characteristics. 
Arranging the scores of all heads across layers yields a two-dimensional importance score matrix, 
which provides a global view of head contributions and serves as the basis for subsequent pruning.

\subsection{Dynamic Pruning}

After evaluating the importance of attention heads, a dynamic pruning strategy is applied to progressively remove 
redundant heads in the multi-head attention mechanism. 
At each training stage, the head with the lowest importance score is pruned, and the pruning ratio is gradually 
increased rather than applied all at once, so as to avoid drastic performance fluctuations.
The pruning ratio at step $t$ is updated as
\begin{equation}
p_{t} = p_{\min} + \left( p_{\max} - p_{\min} \right) \cdot \left( t/T \right)^{n},
\label{eq:ratio_update}
\end{equation}
where $t$ is the current training step, $T$ the total number of steps, and $n$ a hyperparameter controlling the growth rate. Starting from $p_{\min}$, the ratio increases smoothly until the target $p_{\max}$ is reached. Finally, $\left\lfloor p_{\max} \cdot h \right\rfloor$ heads are pruned, where $h$ denotes the total number of attention heads. The overall procedure is summarized in Algorithm~\ref{alg:dynamic-pruning}.

\begin{algorithm}[htbp]
\caption{Dynamic Pruning}
\label{alg:dynamic-pruning}
\begin{algorithmic}[1]  
\REQUIRE Pretrained teacher model $M_T$, training dataset $D$, minimum and maximum pruning ratios $p_{\min}, p_{\max}$, current step $t$;
\ENSURE Pruned student model $M_S$;

\STATE Initialize pruning ratio $p_t = 0$, set total training steps $T$;

\STATE Compute weight norm $w$, attention entropy $H$ using \eqref{eq:attention_weight_norm}\eqref{eq:attention_entropy};
\STATE Compute importance score $S$ using \eqref{eq:importance_score};
\STATE Update pruning ratio $p_t$ according to \eqref{eq:ratio_update};

\STATE Compute number of heads to prune: $h_{prune} = \lfloor p_t \cdot h_{total} \rfloor$;

\IF{$h_{prune} > 0$}
    \STATE Identify $h_{\text{prune}}$ attention heads with the importance score $S$;
    \STATE Prune the selected attention heads;
\ENDIF
    
\STATE Apply pruning and update model structure get student model $M_S$;

\STATE \textbf{return} Pruned student model $M_S$
\end{algorithmic}
\end{algorithm}

\subsection{Recursive Knowledge Distillation}

Although dynamic pruning can significantly reduce the parameter scale and computational cost, some performance 
degradation is unavoidable. 
To address this issue, a recursive knowledge distillation mechanism is introduced, in which pruning and distillation are 
alternated so that the model is gradually compressed while retaining its reasoning capability.

In each iteration, dynamic pruning is first applied to obtain a simplified student model, and the unpruned model is 
regarded as the teacher to guide distillation. 
After each stage, the updated student is adopted as the new teacher for the subsequent round. 
This iterative process continues until the predefined size constraint or performance criterion is satisfied, 
as summarized in Algorithm~\ref{alg:recursive-kd}.

During distillation, the student is optimized with a joint objective consisting of the task loss 
$\mathcal{L}_{\mathrm{task}}$ and the distillation loss $\mathcal{L}_{\mathrm{distill}}$. 
The overall objective is defined as
\begin{equation}
\mathcal{L}_{total} = \lambda_{1} \cdot \mathcal{L}_{\mathrm{distill}} + \lambda_{2} \cdot \mathcal{L}_{\mathrm{task}},
\end{equation}
where $\lambda_{1}$ and $\lambda_{2}$ are balancing coefficients. 
The distillation loss integrates both response-level and feature-level supervision, formulated as
\begin{equation}
\mathcal{L}_{\mathrm{distill}} = \tau^{2} \cdot \mathrm{KL}\!\left(p^{(T)} \,\|\, p^{(S)}\right) + \frac{1}{N}\sum_{l=1}^{N} \mathrm{MSE}\!\left(A_{l}^{(T)}, A_{l}^{(S)}\right),
\end{equation}
where $p^{(T)}$ and $p^{(S)}$ denote the teacher and student output distributions, $\tau$ is the temperature parameter, 
$A_{l}^{(T)}$ and $A_{l}^{(S)}$ are the attention matrices of teacher and student at layer $l$, $N$ is the number of layers, 
$KL$ denotes the Kullback-Leibler divergence, and $MSE$ denotes the mean squared error.

\begin{algorithm}[htbp]
\caption{Recursive Knowledge Distillation}
\label{alg:recursive-kd}
\begin{algorithmic}[1]  
\REQUIRE Pretrained model $M_T$, training dataset $D$;
\ENSURE Lightweight model $M_S$;

\STATE Initialize student model $M'$, set total training steps $T$;

\FOR{$t = 1$ to $T$}
    \STATE Generate student model $M'$ via dynamic pruning using Algorithm \ref{alg:dynamic-pruning};
    \STATE Train student model $M'$ under the guidance of teacher model $M_T$ on dataset $D$;
    \STATE Update teacher model: $M_T \gets M'$;
\ENDFOR

\STATE Lightweight model $M_S \gets M'$;

\STATE \textbf{return} $M_S$
\end{algorithmic}
\end{algorithm}

\section{Experiments}
\label{experiments}

\subsection{Mathematical Reasoning Task Dataset}

\subsubsection{Dataset Overview}

This study evaluates the proposed method on two widely used mathematical reasoning datasets: Math23k and ASDiv-A, 
which together provide diverse test scenarios across languages, reasoning complexity, and arithmetic operations.

\textbf{Math23k} is a large-scale Chinese math word problem dataset released in 2017, 
containing 23,161 questions from primary school curricula (grades 1-6)\cite{wang2017deep}. 
It covers arithmetic, fractions, percentages, geometry, and equation solving, and each sample includes the problem text, 
segmented tokens, the corresponding mathematical expression, and the final answer, making it suitable for evaluating 
expression parsing and construction.

\textbf{ASDiv-A} is an English dataset with 2,305 math word problems, categorized into six difficulty levels\cite{miao2021diverse}. 
Each problem is annotated with solution steps, answers, and formulas, emphasizing multi-step logical reasoning and 
chain-of-thought evaluation.

Together, these datasets complement each other in language and complexity, providing a comprehensive platform to assess 
the efficiency and generalization of lightweight models in mathematical reasoning tasks.

\subsubsection{Dataset Splitting Strategies}

To ensure fair evaluation and generalization, a combination of standardized dataset partitioning and complexity-based 
stratification is employed. Math23k is divided into training, validation and test sets with a 7:1:2 ratio, while ASDiv-A 
follows an 8:1:1 scheme. Stratified sampling is applied to preserve the distribution of problem types across subsets.

Problems are further categorized into three symbolic complexity levels according to the number of operators and nesting depth: 
(1) low complexity, containing one to two operators without nesting; 
(2) medium complexity, involving three to four operators or single-level nesting; and 
(3) high complexity, involving five or more operators or multiple nesting levels. 
This stratification enables a more fine-grained analysis of model performance under varying reasoning difficulties.

\subsection{Experimental Results and Analysis}

\subsubsection{Experimental Setting}

The proposed lightweight method is evaluated on mathematical reasoning tasks using two benchmark models of different scales. 
Both models are based on ATHENA \cite{kim2023athena}, a RoBERTa-based framework optimized for mathematical reasoning 
that integrates neural language modeling with symbolic reasoning through a symbolic attention mechanism. 
ATHENA-base is built on RoBERTa-base with 12 Transformer layers, 768 hidden units, 12 attention heads, and about 125M parameters. 
ATHENA-large is built on RoBERTa-large with 24 layers, 1024 hidden units, 16 attention heads, and about 355M parameters. 
These two configurations serve as complementary baselines for assessing the effectiveness of pruning and distillation 
across different model capacities.

Training uses the AdamW optimizer with an initial learning rate of $5\times 10^{-5}$, weight decay of 0.01, batch size of 32, 
and up to 30 epochs with early stopping. To study compression, pruning ratios of 20\%, 25\%, and 30\% are tested. 
In the distillation stage, the loss combines task and distillation objectives with and temperature $\tau=2$, ensuring effective 
knowledge transfer while maintaining reasoning performance.

\subsubsection{Basic Performance Evaluation}

To comprehensively assess the lightweight effect of dynamic attention head pruning and knowledge distillation, 
extensive experiments were conducted on the Math23k and ASDiv-A datasets. Table~\ref{tab:combined-results} reports 
the performance of ATHENA-base under different pruning ratios after applying the joint optimization method.

\begin{table}[h]
\caption{Performance of pruning-distillation joint optimization on Math23k and ASDiv-A datasets.}
\label{tab:combined-results}
\centering
\begin{adjustbox}{max width=\columnwidth}
\begin{tabular}{@{}c|cccc|cccc@{}}
\toprule
\multirow{2}{*}{Pr.(\%)} & \multicolumn{4}{c|}{(a)Math23k} & \multicolumn{4}{c}{(b)ASDiv-A} \\
\cmidrule(lr){2-5} \cmidrule(lr){6-9}
 & Acc.(\%) & Param↓(\%) & Speed↑(\%) & FLOPs↓(\%) 
 & Acc.(\%) & Param↓(\%) & Speed↑(\%) & FLOPs↓(\%) \\
\midrule
0  & \textbf{84.4}             & 0             & 0             & 0             & \textbf{86.4} & 0    & 0    & 0 \\
20 & 84.2 ({\color{blue}-0.2}) & 12.5          & 18.4          & 12.3          & 86.1 ({\color{blue}-0.3}) & 13.2 & 17.2 & 11.5 \\
25 & 84.0 ({\color{blue}-0.4}) & 15.8          & 22.7          & 15.2          & 85.9 ({\color{blue}-0.5}) & 16.2 & 21.3 & 14.8 \\
30 & 83.7 ({\color{blue}-0.7}) & \textbf{18.7} & \textbf{27.5} & \textbf{17.5} & 85.5 ({\color{blue}-0.9}) & \textbf{19.5} & \textbf{26.5} & \textbf{17.2} \\
\bottomrule
\end{tabular}
\end{adjustbox}
\end{table}

On the Math23k dataset, as shown in Table~\ref{tab:combined-results}(a), increasing the pruning ratio substantially 
reduces parameters, memory, and computation, while accuracy decreases only slightly. At a 30\% pruning ratio, 
accuracy drops from 84.4\% to 83.7\% (-0.7\%), accompanied by an 18.7\% parameter reduction and a 27.5\% speedup. 
These results confirm that the proposed method achieves significant efficiency gains while effectively maintaining 
reasoning performance.

The method was further evaluated on the ASDiv-A dataset to assess its generalization ability. 
As shown in Table~\ref{tab:combined-results}(b), the method also performs well on English mathematical reasoning. 
At a 25\% pruning ratio, accuracy decreases marginally from 86.4\% to 85.9\% (-0.5\%), 
while parameters are reduced by 16.2\% and inference speed is improved by 21.3\%, 
demonstrating that the joint optimization consistently achieves significant efficiency gains with minimal accuracy loss across datasets.

\subsubsection{Effects of Different Model Sizes}

To examine the applicability of the proposed method across model scales, 
the pruning-distillation optimization was applied to ATHENA models based on both RoBERTa-base and RoBERTa-large. 
The results on Math23k and ASDiv-A are summarized in Table~\ref{tab:model-scale}.

\begin{table}[h]
\caption{Performance of pruning-distillation joint optimization on models of different scales.}
\label{tab:model-scale}
\centering
\begin{adjustbox}{max width=\columnwidth}
\scriptsize
\begin{tabular}{@{}c|cccc|cccc@{}}
\toprule
\multirow{2}{*}{Model} & \multicolumn{4}{c|}{(a)Math23k} & \multicolumn{4}{c}{(b)ASDiv-A} \\
\cmidrule(lr){2-5} \cmidrule(lr){6-9}
& Pr.(\%) & Acc.(\%) & Param↓(\%) & Speed↑(\%) & Pr.(\%) & Acc.(\%) & Param↓(\%) & Speed↑(\%) \\
\midrule
\multirow{2}{*}{ATHENA-base}
& 0  & \textbf{84.4} & 0    & 0    & 0  & \textbf{86.4} & 0    & 0 \\
& 25 & \underline{84.0} & \textbf{15.8} & \textbf{22.7} & 25 & \underline{85.9} & \textbf{16.2} & \textbf{21.3} \\
\midrule
\multirow{2}{*}{ATHENA-large}
& 0  & \textbf{86.5} & 0    & 0    & 0  & \textbf{91.0} & 0    & 0 \\
& 25 & \underline{86.0} & \textbf{17.4} & \textbf{23.8} & 25 & \underline{90.4} & \textbf{18.1} & \textbf{22.5} \\
\bottomrule
\end{tabular}
\end{adjustbox}
\end{table}

On Math23k, ATHENA-large also benefits from the proposed method, as shown in Table~\ref{tab:model-scale}(a). 
At a 25\% pruning ratio, accuracy decreases by only 0.5\%, while parameters are reduced by 17.4\% and inference speed is improved by 23.8\%. 
For comparison, ATHENA-base shows a 0.4\% accuracy drop under the same setting, with 15.8\% parameter reduction and 22.7\% speedup. 
This indicates that larger models contain more redundancy in attention heads, making the optimization more effective.

A similar trend is observed on ASDiv-A. As shown in Table~\ref{tab:model-scale}(b), ATHENA-large accuracy decreases from 
91.0\% to 90.4\% (-0.6\%) at 25\% pruning, while parameters are reduced by 18.1\% and inference speed is improved by 22.5\%. 
These results further confirm that the proposed method consistently delivers efficiency gains with minimal accuracy loss 
across different model sizes and datasets.

\subsubsection{Impact of the Hyperparameter $\alpha$}

To explore the relative contribution of weight norm and attention entropy in the attention head importance score, 
sensitivity experiments were conducted on the hyperparameter $\alpha$. As shown in Eq.~\ref{eq:importance_score}, 
$\alpha \in [0,1]$ controls their weighting: $\alpha=0$ uses only entropy, $\alpha=1$ only norm, and intermediate values 
combine both. Table~\ref{tab:analysis}(a) reports results on ATHENA-base with Math23k at a 25\% pruning ratio.

\begin{table}[htbp]
\caption{Performance of pruning-distillation joint optimization under different $\alpha$ values and problem complexity levels (Math23k, 25\% pruning).}
\label{tab:analysis}
\centering
\begin{subtable}[t]{0.49\columnwidth}
\centering
\scriptsize
\begin{tabular}{ccccc}
\toprule
$\alpha$ & Acc.(\%) & Param↓(\%) & Speed↑(\%) & FLOPs↓(\%) \\
\midrule
0.0 & 83.1 & 15.8 & 22.7 & 15.2 \\
0.3 & \underline{83.8} & 15.8 & 22.7 & 15.2 \\
0.5 & \textbf{84.0} & 15.8 & 22.7 & 15.2 \\
0.7 & 83.5 & 15.8 & 22.7 & 15.2 \\
1.0 & 71.2 & 15.8 & 22.7 & 15.2 \\
\bottomrule
\end{tabular}
\caption{Impact of different $\alpha$ values.}
\end{subtable}
\hfill
\begin{subtable}[t]{0.49\columnwidth}
\centering
\scriptsize
\begin{tabular}{cccc}
\toprule
\makecell{Problem \\ Complexity} & \makecell{Baseline \\ Acc.(\%)} &  \makecell{Optimized \\ Acc.(\%)} &  \makecell{$\Delta$Acc.\\(\%)} \\
\midrule
Simple  & 89.2 & 88.9 & \color{blue}-0.3 \\
Medium  & 84.7 & 84.2 & \color{blue}-0.5 \\
Complex & 79.3 & 78.6 & \color{blue}-0.7 \\
All     & 84.4 & 84.0 & \color{blue}-0.4 \\
\bottomrule
\end{tabular}
\caption{Impact of problem complexity.}
\end{subtable}
\end{table}

The experimental results reveal three key observations. 
First, the best performance is achieved at $\alpha=0.5$, where the accuracy reaches 84.0\%, indicating that weight norm 
and attention entropy contribute comparably and should be balanced. 
Second, using only entropy ($\alpha=0$) yields an accuracy of 83.1\%, which is markedly higher than the 71.2\% obtained 
when relying solely on norm ($\alpha=1$), showing that entropy provides more informative guidance by reflecting the 
richness of attention distributions, whereas norm merely measures parameter scale. 
Third, when $\alpha$ lies between 0.3 and 0.7, accuracy remains above 83.5\%, but it drops sharply as $\alpha$ 
approaches 1, suggesting that over-reliance on norm misguides pruning.

Therefore, all experiments in this paper are conducted with $\alpha=0.5$, providing a robust balance between entropy 
and norm for reliable head evaluation.

\subsubsection{Problem Complexity Hierarchical Analysis}

To better understand the effect of the proposed method on problems of different complexity, the Math23k test set was 
divided into simple, medium, and complex categories, and model performance was evaluated before and after optimization. 
Table~\ref{tab:analysis}(b) reports the results of ATHENA-base under a 25\% pruning ratio.

The results show that the proposed method maintains stable performance across different complexity levels. 
For simple problems, accuracy decreases by only 0.3\%; for complex problems, the drop is 0.7\%, slightly higher than the 
overall average but still acceptable. This demonstrates that the method remains effective even for challenging reasoning 
tasks, offering a practical solution in resource-constrained settings.

\subsection{Ablation Study}

To systematically evaluate the contribution of each component, ablation experiments were conducted on ATHENA-base with 
Math23k at a pruning ratio of 25\%. The results are reported in Table~\ref{tab:ablation}.

\begin{table}[h]
\caption{Ablation study of optimization components (ATHENA-base, Math23k, 25\% pruning).}
\label{tab:ablation}
\centering
\scriptsize
\begin{adjustbox}{max width=\columnwidth}
\begin{tabular}{@{}c|c|ccccc|cccc|c@{}}
\toprule
\multirow{4}{*}{Strategy} & \multirow{4}{*}{\makecell{Baseline}} 
& \multicolumn{5}{c|}{w/o Recursive Knowledge Distillation} 
& \multicolumn{4}{c|}{w/o Dynamic Pruning} 
& \multirow{4}{*}{\makecell{Ours}} \\ 
\cmidrule(lr){3-7} \cmidrule(lr){8-11}
& & \makecell{pruning \\ only} 
  & \makecell{w/o \\ entropy} 
  & \makecell{w/o \\ norm} 
  & \makecell{w. resp. \\ distill} 
  & \makecell{w. feat. \\ distill} 
  & \makecell{random \\ pruning} 
  & \makecell{global thr. \\ pruning} 
  & \makecell{attention \\ vis. pruning} 
  & \makecell{distill \\ only}
  & \\ 
\midrule
Acc.(\%) 
& \underline{84.4}
& 83.6 & 71.2 & 83.1 & 83.8 & 83.9 
& 81.5 & 82.9 & 83.6 & \textbf{84.5} 
& {\color{red}84.0} \\
Param↓(\%) 
& 0.0 
& 15.8 & 15.8 & 15.8 & 15.8 & 15.8 
& 15.8 & 15.8 & 15.8 & 0.0 
& \textbf{15.8} \\
Speed↑(\%) 
& 0.0 
& 22.7 & 22.7 & 22.7 & 22.7 & 22.7 
& 22.7 & 22.7 & 22.7 & 0.0 
& \textbf{22.7} \\
Memory↓(\%) 
& 0.0 
& 11.6 & 11.6 & 11.6 & 11.6 & 11.6 
& 11.6 & 11.6 & 11.6 & 0.0 
& \textbf{11.6} \\
\bottomrule
\end{tabular}
\end{adjustbox}
\end{table}

Without recursive knowledge distillation, different pruning criteria were examined. Using both weight norm and attention 
entropy achieves 83.6\% accuracy, while removing norm slightly reduces it to 83.1\%. In contrast, removing entropy causes 
a drastic drop to 71.2\% (13.2\% below baseline), indicating that entropy is indispensable and that the two metrics are 
complementary in evaluating head importance.
When distillation strategies were introduced, response distillation (resp. distill) improves accuracy to 83.8\% and 
feature distillation (feat. distill) to 83.9\%. However, both remain inferior to the proposed joint optimization, 
which reaches 84.0\% while simultaneously reducing parameters, computation, and memory. 

Without dynamic pruning, mainstream pruning strategies were compared. Random pruning yields 81.5\%,
global threshold (global thr.) pruning 82.9\%, and attention visualization (attention vis.) pruning 83.6\%. 
Compared with our method, the accuracy gaps are 2.5\%, 1.1\%, and 0.4\%, respectively.
A distillation-only setting achieves 84.5\% accuracy but retains the same parameter size and memory as the baseline, 
thus offering no efficiency benefits.

Overall, the proposed method achieves 84.0\% accuracy with only 0.4\% loss relative to baseline, while reducing 
parameters by about 16\%, improving inference speed by over 22\%, and lowering memory by around 12\%. These results 
demonstrate that dynamic pruning combined with recursive knowledge distillation provides the most effective trade-off 
between accuracy and efficiency.

\section{Conclusion}
\label{conclusion}

In this paper, a lightweighting method for large language models in mathematical reasoning tasks was investigated to 
address the challenges of high computational cost and storage demand. 
The proposed approach integrates dynamic pruning with recursive knowledge distillation. 
Attention head importance is evaluated through a joint metric combining weight norm and attention entropy, 
allowing redundant heads to be pruned adaptively during training. To mitigate the performance loss caused by pruning, 
knowledge distillation is applied in a recursive manner, where the unpruned model first guides the pruned student, 
and the student then iteratively serves as the new teacher. 
This strategy achieves structural compression while preserving reasoning accuracy. 
On Math23k, for example, a 30\% pruning ratio reduced parameters by 18.7\%, improved inference speed by 27.5\%, 
and caused only a 0.7\% accuracy drop. 
Consistent results across datasets, model scales, and problem complexities demonstrate the practicality of the method 
for efficient deployment in resource-constrained environments.

The significance of this study is both theoretical and practical. Theoretically, a multi-dimensional evaluation mechanism 
based on weight norm and attention entropy was proposed, offering a new perspective for optimizing 
multi-head attention and revealing the specialization phenomenon of attention heads in mathematical reasoning. 
Practically, the method reduces resource requirements while preserving high performance, enabling efficient deployment 
of large models in constrained environments.

Future research may focus on validating the method in real teaching scenarios, adapting and optimizing it for diverse 
hardware environments, and applying more advanced causal analysis to identify key attention heads. 
Further exploration of integrated lightweighting strategies, such as combining pruning with quantization or parameter 
sharing, as well as evaluation across multiple NLP tasks and multi-task settings, will also be valuable.

\bibliographystyle{splncs04}
\bibliography{clean}

@inproceedings{vaswani2017attention,
  author    = {Vaswani, A. and Shazeer, N. and Parmar, N.},
  booktitle = {Advances in Neural Information Processing Systems},
  title     = {Attention is all you need},
  volume    = {30},
  year      = {2017}
}

@article{brown2020language,
  author  = {Brown, Tom and Mann, Benjamin and Ryder, Nick and
             Subbiah, Melanie and Kaplan, Jared D and
             Dhariwal, Prafulla and Neelakantan, Arvind and
             Shyam, Pranav and Sastry, Girish and Askell, Amanda and
             others},
  journal = {Advances in neural information processing systems},
  pages   = {1877--1901},
  title   = {Language models are few-shot learners},
  volume  = {33},
  year    = {2020}
}

@article{kim2023athena,
  author  = {Kim, M. and Kang, D. and Lee, S.},
  journal = {Conference on Empirical Methods in Natural Language
             Processing},
  pages   = {4932--4947},
  title   = {ATHENA: Mathematical reasoning with thought
             expansion},
  year    = {2023}
}

@article{liang2021mwp,
  author  = {Liang, Z. and Zhang, J. and Wang, L.},
  journal = {arXiv preprint arXiv:2107.13435},
  title   = {MWP-BERT: Numeracy-augmented pre-training for math
             word problem solving},
  year    = {2021}
}

@inproceedings{zong2023solving,
  author    = {Zong, M. and Krishnamachari, B.},
  booktitle = {Proceedings of the AAAI Conference on Artificial
               Intelligence},
  number    = {13},
  pages     = {15972--15979},
  title     = {Solving math word problems concerning systems of
               equations with GPT-3},
  volume    = {37},
  year      = {2023}
}

@article{jin2024moh,
  author  = {Jin, P. and Zhu, B. and Yuan, L.},
  journal = {arXiv preprint arXiv:2410.11842},
  title   = {MoH: Multi-head attention as mixture-of-head
             attention},
  year    = {2024}
}

@book{feigenbaum1963computers,
  author    = {Feigenbaum, E. A. and Feldman, J.},
  publisher = {McGraw-Hill},
  title     = {Computers and thought},
  year      = {1963}
}

@article{ling2017program,
  author  = {Ling, Wang and Yogatama, Dani and Dyer, Chris and
             Blunsom, Phil},
  journal = {Proceedings of the 55th Annual Meeting of the
             Association for Computational Linguistics},
  pages   = {158--167},
  title   = {Program induction by rationale generation: Learning
             to solve and explain algebraic word problems},
  year    = {2017}
}

@article{wang2017deep,
  author  = {Wang, Yan and Liu, Xiaojiang and Shi, Shuming},
  journal = {Proceedings of the 2017 Conference on Empirical
             Methods in Natural Language Processing},
  pages   = {845--854},
  title   = {Deep neural solver for math word problems},
  year    = {2017}
}

@article{bengio1994learning,
  author  = {Bengio, Yoshua and Simard, Patrice and
             Frasconi, Paolo},
  journal = {IEEE transactions on neural networks},
  number  = {2},
  pages   = {157--166},
  title   = {Learning long-term dependencies with gradient descent
             is difficult},
  volume  = {5},
  year    = {1994}
}

@article{zhang2020graph,
  author  = {Zhang, J. and Wang, L. and Lee, R.},
  journal = {Association for Computational Linguistics},
  pages   = {3928--3937},
  title   = {Graph-to-tree learning for solving math word
             problems},
  year    = {2020}
}

@article{hoefler2021sparsity,
  author  = {Hoefler, T. and Alistarh, D. and Ben-Nun, T.},
  journal = {Journal of Machine Learning Research},
  number  = {241},
  pages   = {1--124},
  title   = {Sparsity in deep learning: Pruning and growth for
             efficient inference and training in neural networks},
  volume  = {22},
  year    = {2021}
}

@inproceedings{han2015learning,
  author    = {Han, S. and Pool, J. and Tran, J.},
  booktitle = {Advances in Neural Information Processing Systems},
  title     = {Learning both weights and connections for efficient
               neural network},
  volume    = {28},
  year      = {2015}
}

@article{sanh2019distilbert,
  author  = {Sanh, V. and Debut, L. and Chaumond, J.},
  journal = {arXiv preprint arXiv:1910.01108},
  title   = {DistilBERT, a distilled version of BERT: smaller,
             faster, cheaper and lighter},
  year    = {2019}
}

@article{gou2021knowledge,
  author  = {Gou, J. and Yu, B. and Maybank, S. J.},
  journal = {International Journal of Computer Vision},
  number  = {6},
  pages   = {1789--1819},
  title   = {Knowledge distillation: A survey},
  volume  = {129},
  year    = {2021}
}

@article{2024-40694,
  author  = {Haiwei Pan and Fengming Yu and Kejia Zhang and
             Haiyan Lan and Qingyu Meng and Zhe Li},
  journal = {Journal of Computer Research and Development},
  number  = {},
  pages   = {1-33},
  title   = {Knowledge Distillation in Visual Algorithms: A
             Survey},
  volume  = {62},
  year    = {2025}
}

@article{frankle2020lottery,
  author  = {Frankle, J. and Carbin, M.},
  journal = {International Conference on Learning Representations},
  title   = {The lottery ticket hypothesis: Finding sparse,
             trainable neural networks},
  year    = {2020}
}

@article{gao2018dynamic,
  author  = {Gao, X. and Zhao, Y. and Dudziak, L.},
  journal = {International Conference on Learning Representations},
  title   = {Dynamic channel pruning: Feature boosting and
             suppression},
  year    = {2018}
}

@inproceedings{fang2023depgraph,
  author    = {Fang, G. and Ma, X. and Song, M.},
  booktitle = {Proceedings of the IEEE/CVF Conference on Computer
               Vision and Pattern Recognition},
  pages     = {16091--16101},
  title     = {DepGraph: Towards any structural pruning},
  year      = {2023}
}

@inproceedings{frantar2023sparsegpt,
  author       = {Frantar, E. and Alistarh, D.},
  booktitle    = {International Conference on Machine Learning},
  organization = {PMLR},
  pages        = {10323--10337},
  title        = {SparseGPT: Massive language models can be accurately
                  pruned in one-shot},
  year         = {2023}
}

@article{hinton2015distilling,
  author  = {Hinton, G. and Vinyals, O. and Dean, J.},
  journal = {arXiv preprint arXiv:1503.02531},
  title   = {Distilling the knowledge in a neural network},
  year    = {2015}
}

@inproceedings{DBLP:conf/iccv/SonNCH21,
  author    = {Wonchul Son and Jaemin Na and Junyong Choi and
               Wonjun Hwang},
  booktitle = {2021 {IEEE/CVF} International Conference on Computer
               Vision},
  pages     = {9375--9384},
  publisher = {{IEEE}},
  title     = {Densely Guided Knowledge Distillation using Multiple
               Teacher Assistants},
  year      = {2021}
}

@inproceedings{DBLP:journals/corr/RomeroBKCGB14,
  author    = {Adriana Romero and Nicolas Ballas and
               Samira Ebrahimi Kahou and Antoine Chassang and
               Carlo Gatta and Yoshua Bengio},
  booktitle = {3rd International Conference on Learning
               Representations},
  title     = {FitNets: Hints for Thin Deep Nets},
  year      = {2015}
}

@inproceedings{DBLP:conf/iccv/LaoSLLY23,
  author    = {Shanshan Lao and Guanglu Song and Boxiao Liu and
               Yu Liu and Yujiu Yang},
  booktitle = {{IEEE/CVF} International Conference on Computer
               Vision},
  pages     = {6361--6370},
  publisher = {{IEEE}},
  title     = {Masked Autoencoders Are Stronger Knowledge
               Distillers},
  year      = {2023}
}

@inproceedings{DBLP:conf/cvpr/YimJBK17,
  author    = {Junho Yim and Donggyu Joo and Ji{-}Hoon Bae and
               Junmo Kim},
  booktitle = {{IEEE/CVF} Conference on Computer Vision and Pattern
               Recognition},
  pages     = {7130--7138},
  publisher = {{IEEE}},
  title     = {A Gift from Knowledge Distillation: Fast
               Optimization, Network Minimization and Transfer
               Learning},
  year      = {2017}
}

@inproceedings{DBLP:conf/iclr/XuFZXWDX022,
  author    = {Haohang Xu and Jiemin Fang and Xiaopeng Zhang and
               Lingxi Xie and Xinggang Wang and Wenrui Dai and
               Hongkai Xiong and Qi Tian},
  booktitle = {The Tenth International Conference on Learning
               Representations},
  publisher = {OpenReview.net},
  title     = {Bag of Instances Aggregation Boosts Self-supervised
               Distillation},
  year      = {2022}
}

@article{miao2021diverse,
  author  = {Miao, Shen-Yun and Liang, Chao-Chun and Su, Keh-Yih},
  journal = {arXiv preprint arXiv:2106.15772},
  title   = {A diverse corpus for evaluating and developing
             English math word problem solvers},
  year    = {2021}
}

\end{document}